\documentclass[10pt,twocolumn,letterpaper]{article}

\usepackage{cvpr}
\usepackage{times}
\usepackage{epsfig}
\usepackage{graphicx}
\usepackage{amsmath}
\usepackage{amssymb}
\usepackage{stmaryrd}
\usepackage{bm}
\renewcommand\vec[1]{\overset{{}_{\shortrightarrow}}{#1}}
\newcommand\cev[1]{\overset{{}_{\shortleftarrow}}{#1}}

\usepackage[breaklinks=true,bookmarks=false]{hyperref}

\cvprfinalcopy 

\def\cvprPaperID{****} 

\begin{document}

\title{Learning to Compose Dynamic Tree Structures for Visual Contexts}

\author{Kaihua Tang$^1$, Hanwang Zhang$^1$, Baoyuan Wu$^2$, Wenhan Luo$^2$, Wei Liu$^2$ \\
$^1$ Nanyang Technological University \\
$^2$ Tencent AI Lab \\
$\textit{kaihua001@e.ntu.edu.sg}$, $\textit{hanwangzhang@ntu.edu.sg}$, \\
$\textit{\{wubaoyuan1987, whluo.china\}@gmail.com}$, $\textit{wl2223@columbia.edu}$
}

\maketitle

\begin{abstract}
We propose to compose dynamic tree structures that place the objects in an image into a visual context, helping visual reasoning tasks such as scene graph generation and visual Q\&A.
Our visual context tree model, dubbed \textsc{VCTree}, has two key advantages over existing structured object representations including chains and fully-connected graphs: 1) The efficient and expressive binary tree encodes the inherent parallel/hierarchical relationships among objects, \eg, ``clothes'' and ``pants'' are usually co-occur and belong to ``person''; 2) the dynamic structure varies from image to image and task to task, allowing more content-/task-specific message passing among objects. To construct a \textsc{VCTree}, we design a score function that calculates the task-dependent validity between each object pair, and the tree is the binary version of the maximum spanning tree from the score matrix. Then, visual contexts are encoded by bidirectional TreeLSTM and decoded by task-specific models. We develop a hybrid learning procedure which integrates end-task supervised learning and the tree structure reinforcement learning, where the former's evaluation result serves as a self-critic for the latter's structure exploration.  Experimental results on two benchmarks, which require reasoning over contexts: Visual Genome for scene graph generation and VQA2.0 for visual Q\&A, show that \textsc{VCTree} outperforms state-of-the-art results while discovering interpretable visual context structures.
\end{abstract}

\section{Introduction}

\begin{figure}[t]
   \begin{minipage}[b]{1.0\linewidth}
   \centerline{\includegraphics[width=85mm]{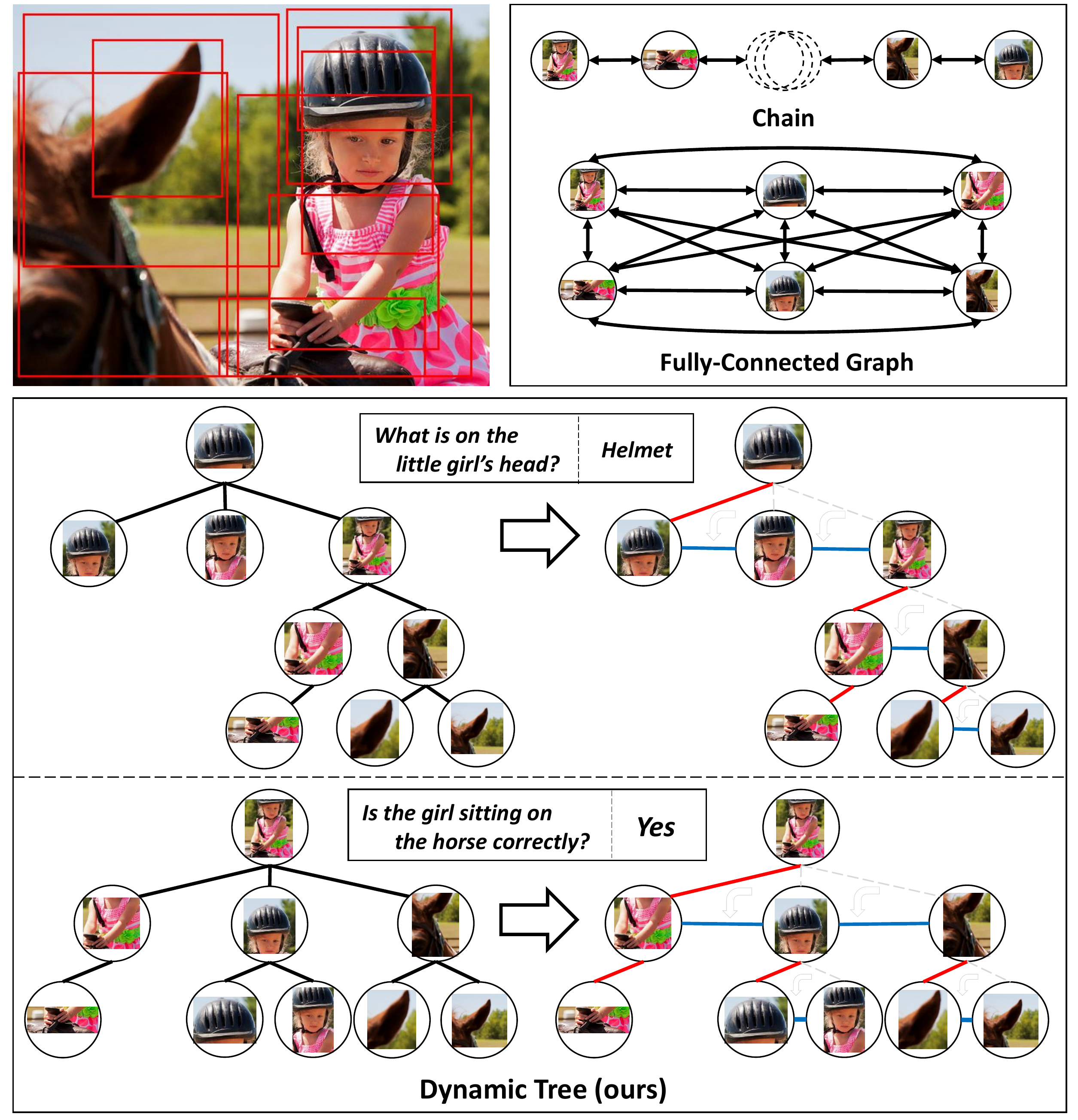}}
   \end{minipage}
   \caption{Illustrations of different object-level visual context structures: chains~\cite{zellers2017neural}, fully-connected graphs~\cite{xu2017scene}, and dynamic tree structures constructed by the proposed \textsc{VCTree}. For the purpose of efficient context encoding by using TreeLSTM~\cite{tai2015improved}, we transform the multi-branch trees (left) to the equivalent left-child right-sibling binary trees~\cite{Cormen2001}, where the left branches (red) indicate the hierarchical relations and right branches (blue) indicate the parallel relations. The key advantages of \textsc{VCTree} over chains and graphs are hierarchical, dynamic, and efficient.}
   \vspace{-0.1in}
   \label{fig:1} 
\end{figure}


Objects are not alone. They are placed in the visual context: a coherent object configuration attributed to the fact that they co-vary with each other. Extensive studies in cognitive science show that our brains inherently exploit visual contexts to understand cluttered visual scenes comprehensively~\cite{bar2004visual, biederman1982scene, oliva2007role}. For example, even the girl's leg and the horse are not fully observed in Figure~\ref{fig:1}, we can still infer ``girl riding horse''. Inspired by this, modeling visual contexts is also indispensable in many modern computer vision systems. For example, state-of-the-art CNN architectures capture the context by convolutions of various receptive fields and encode it into multi-scale feature map pyramid~\cite{chen2018deeplab, lin2017feature, zhao2015saliency}. Such pixel-level visual context (or local context~\cite{divvala2009empirical}) arguably plays one of the key roles in closing the performance gap of the ``mid-level'' vision between humans and machines, such as R-CNN based object detection~\cite{lin2017feature, liu2016ssd, ren2015faster}, instance segmentation~\cite{he2017mask, pinheiro2016learning}, and FCN based semantic segmentation~\cite{chen2018deeplab, chen2017rethinking, yu2015multi}.

Modeling visual contexts \emph{explicitly} on the object-level has also been shown effective in ``high-level'' vision tasks such as image captioning~\cite{Yao_2018_ECCV} and visual Q\&A~\cite{Teney_2017_CVPR}. In fact, the visual context serves as a powerful inductive bias that connects objects in a particular layout for high-level reasoning~\cite{li2017scene, liu2018structure, Teney_2017_CVPR, Yao_2018_ECCV}. For example, the spatial layout of ``person'' on ``horse'' is useful for determining the relationship ``ride'', which is in turn informative to localize the ``person'' if we want to answer ``who is riding on the horse?''. However, those works assume that the context is a scene graph, whose detection \textit{per se} is a high-level task and not yet reliable. Without high-quality scene graphs, we have to use a prior layout structure. As shown in Figure~\ref{fig:1}, two popular structures are chains~\cite{zellers2017neural} and fully-connected graphs~\cite{Chen_2018_CVPR, dai2017detecting, li2018factorizable, xu2017scene, yin2018zoom}, where the context is encoded by sequential models such as bidirectional LSTM~\cite{hochreiter1997long} for chains and CRF-RNN~\cite{zheng2015conditional} for graphs. 

However, these two prior structures are sub-optimal. First, chains are oversimplified and may only capture simple spatial information or co-occurrence bias; though fully-connected graphs are complete, they lack the discrimination between hierarchical relations, \eg, ``helmet affiliated to head'', and parallel relations, \eg,  ``girl on horse''; in addition, dense connections could also lead to message passing saturation in the subsequent context encoding~\cite{xu2017scene}. Second, visual contexts are inherently content-/task-driven, \eg, the object layouts should vary from content to content, question to question. Therefore, fixed chains and graphs are incompatible with the dynamic nature of visual contexts~\cite{watanabe1998task}. 

In this paper, we propose a model dubbed \textsc{VCTree}, pioneering to compose dynamic tree structures for encoding object-level visual context for high-level visual reasoning tasks, such as scene graph generation (SGG) and visual Q\&A (VQA). Given a set of object proposals in an image (\eg, obtained from Faster-RCNN~\cite{ren2015faster}), we maintain a trainable task-specific score matrix of the objects, where each entry indicates the contextual validity of the pairwise objects. Then, a maximum spanning tree can be trimmed from the score matrix, \eg, the multi-branch trees shown in Figure~\ref{fig:1}. This dynamic structure represents a ``hard'' hierarchical layout bias of what objects should gain more contextual information from others, \eg, objects on the person's head are most informative given the question ``what on the little girl's head?''; while the whole person's body is more important given the question ``Is the girl sitting on the horse correctly?''. To avoid the saturation issue caused by the densely connected arbitrary number of children, we further morph the multi-branch trees to the equivalent left-child right-sibling binary trees~\cite{Cormen2001}, where the left branches (red) indicate the hierarchical relations and right branches (blue) indicate the parallel relations, then use TreeLSTM~\cite{tai2015improved} to encode the context.

As the above \textsc{VCTree} construction is in a discrete and non-differentiable nature, we develop a hybrid learning strategy using REINFORCE~\cite{hu2017learning, Rennie_2017_CVPR, williams1992simple} for tree structure exploration and supervised learning for context encoding and its subsequent tasks. In particular, the evaluation result (Recall for SGG and Accuracy for VQA) from supervised task can be exploited as a “critic” function that guide the ``action'' of tree construction.  We evaluate \textsc{VCTree} on two benchmarks: Visual Genome~\cite{krishna2017visual} for SGG and VQA2.0~\cite{goyal2017making} for VQA. For SGG, we achieve a new state-of-the-art on all three standard tasks, \ie, Scene Graph Generation, Scene Graph Classification, and Predicate Classification; for VQA, we achieve competitive results on single model performances. In particular, \textsc{VCTree} helps high-level vision models fight against the dataset bias. For example, we achieve 4.1\% absolute gain in proposed Mean Recall@100 metric of Predicate Classification than MOTIFS~\cite{zellers2017neural}, and observe higher improvement in VQA2.0 balanced pair subset~\cite{Teney_2018_CVPR} than normal validation set. Qualitative results also show that \textsc{VCTree} composes interpretable structures.

\section{Related Work}
\noindent\textbf{Visual Context Structures}. Despite the consensus on the value of visual contexts, existing context models are diversified into a variety of implicit or explicit approaches. Implicit models directly encode surrounding pixels into multi-scale feature maps, \eg, dilated convolution~\cite{yu2015multi} presents a efficient way to increase receptive field, applicable in various dense prediction tasks~\cite{chen2018deeplab, chen2017rethinking}; feature pyramid structure~\cite{lin2017feature} combines low-resolution contextual features with high-resolution detailed features, facilitating object detection with rich semantics. Explicit models incorporate contextual cues through object connections. However, such methods~\cite{li2018factorizable, xu2017scene, zellers2017neural} group objects into fixed layouts, \ie, chains or graphs.

\noindent\textbf{Learning to Compose Structures}. Learning to compose structures is becoming popular in NLP for sentence representation, \eg, Cho~\etal~\cite{cho2014properties} applied a gated recursive convolutional neural network (grConv) to control the bottom-up feature flow for a dynamic structure; Choi~\etal~\cite{choi2018learning} combines TreeLSTM with Gumbel-Softmax, allowing task-specific tree structures automatically learned from plain text. Yet, only few works compose visual structures for images. Conventional approaches construct a statistical dependency graph/tree for the entire dataset based on object categories~\cite{choi2012tree} or exemplars~\cite{malisiewicz2009beyond}. Those statistical methods cannot put per-image objects in a context as a whole to reason over content-/task-specific fashion. Socher~\etal~\cite{socher2011parsing} constructed a bottom-up tree structure to parse images; however, their tree structure learning is supervised while ours is reinforced, which does not require tree ground-truth. 


\begin{figure*}[t]
   \begin{minipage}[b]{1.0\linewidth}
   \centerline{\includegraphics[width=180mm]{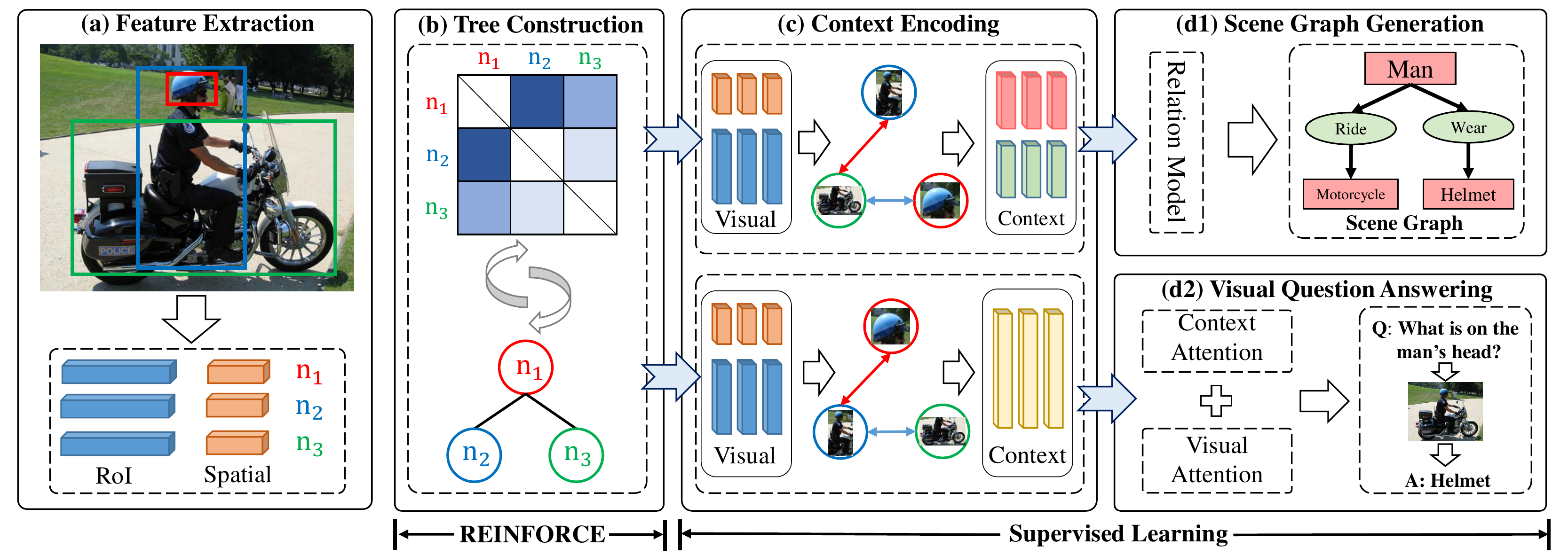}}
   \end{minipage}
   \caption{The framework of the proposed \textsc{VCTree} model. We extract visual features from proposals and construct a dynamic \textsc{VCTree} using the learnable score matrix. The tree structure is used to encode the object-level visual context, which will be decoded for each specific end-task. Parameters in stages (c)\&(d) are trained by supervised learning, while those in stage (b) are using REINFORCE with a self-critic baseline.}
   \vspace{-0.1in}
  \label{fig:2} 
\end{figure*}

\noindent\textbf{Visual Reasoning Tasks}. Scene Graph Generation (SGG) task is derived from Visual Relationship Detection (VRD). Early work on VRD~\cite{lu2016visual} treats objects as isolated individuals, while SGG considers each image as a whole. Along with the widely used message passing mechanism~\cite{xu2017scene}, a variety of context models~\cite{li2018factorizable, li2017scene,newell2017pixels, Yang_2018_ECCV} have been exploited in SGG to fine-tune local predictions through rich global contexts, making it the best competition field for different contextual models. Visual Question Answering (VQA) as a high-level task bridges the gap between computer vision and natural language processing. State-of-the-art VQA models~\cite{anderson2018bottom, bai2018deep, Teney_2018_CVPR} rely on bag-of-object visual attentions which can be considered as a trivial context structure. However, we propose to learn a tree context structure that is dynamic to visual content and questions.

\section{Approach}
\label{section3}
As illustrated in Figure~\ref{fig:2}, our \textsc{VCTree} model can be summarized into the following four steps. (a) We adopt Faster-RCNN to detect object proposals~\cite{ren2015faster}. The visual feature of each proposal $i$ is presented as $\bm{x}_i$, concatenating a RoIAlign feature~\cite{he2017mask} $\bm{v}_i\in\mathbb{R}^{2048}$ and spatial feature $\bm{b}_i \in \mathbb{R}^8$, where 8 elements indicate the bounding box coordinates $(x_1,y_1,x_2,y_2)$, center $(\frac{x_1+x_2}{2},\frac{y_1+y_2}{2})$, and size $(x_2-x_1,y_2-y_1)$, respectively. Note that the visual feature $\bm{x}_i$ is not limited to bounding box; segment feature from instance segmentations~\cite{he2017mask} or panoptic segmentations~\cite{kirillov2018panoptic} could also be alternatives. (b) In Section~\ref{subsection:3_1}, a learnable matrix will be introduced to construct \textsc{VCTree}. Moreover, since the \textsc{VCTree} construction is discrete in nature and the score matrix is non-differentiable from the loss of end-task, we develop a hybrid learning strategy in Section~\ref{subsection:3_5}. (c) In Section~\ref{subsection:3_2}, we employ Bidirectional Tree LSTM (BiTreeLSTM) to encode the contextual cues using the constructed \textsc{VCTree}. (d) The encoded contexts will be decoded for each specific end-task detailed in Section~\ref{subsection:3_3} and Section~\ref{subsection:3_4}.

\subsection{\textsc{VCTree} Construction}
\label{subsection:3_1}
\textsc{VCTree} construction aims to learn a score matrix $\bm{S}$, which approximates the task-dependent validity between each object pair. Two principles guide the formulation of this matrix: 1) inherent object correlations should be maintained, \eg, ``man wears helmet'' in Figure~\ref{fig:2}; (2) task related object pair has higher score than irrelevant ones, \eg, given question ``what is on the man's head?'', ``man-helmet'' pair should be more important than ``man-motorcycle'' and ``helmet-motorcycle'' pairs. Therefore, we define each element of $\bm{S}$ as the product of the object correlation $f(\bm{x}_i,\bm{x}_j)$ and the pairwise task-dependency $g(\bm{x}_i,\bm{x}_j, \bm{q})$:
\begin{equation}
\left\{
    \begin{array}{lr}
    \bm{S}_{ij} = f(\bm{x}_i,\bm{x}_j)\cdot g(\bm{x}_i,\bm{x}_j, \bm{q}), \\
    f(\bm{x}_i,\bm{x}_j) =\sigma\left(\textnormal{MLP}(\bm{x}_i,\bm{x}_j)\right), \\
    g(\bm{x}_i,\bm{x}_j, \bm{q})=\sigma(h(\bm{x}_i,\bm{q}))\cdot \sigma(h(\bm{x}_j,\bm{q})), \\
    \end{array}
\right.
\label{eq:1}
\end{equation}
where $\sigma(\cdot)$ is the sigmoid function; $\bm{q}$ is the task feature, \eg, the question feature encoded by GRU in VQA; MLP is a multi-layer perceptron; $h(\bm{x}_i,\bm{q})$ is the object-task correlation in VQA, which will be introduced later in Section~\ref{subsection:3_4}. In SGG, the entire $g(\bm{x}_i,\bm{x}_j, \bm{q})$ is set to $1$, as we assume that each object pair contributes equally without the question prior. We pretrain $f(\bm{x}_i,\bm{x}_j)$ on Visual Genome~\cite{krishna2017visual} for a reasonable binary prior if two objects are related. Yet, such a pretrained model is not perfect due to the lack of coherent graph-level constraint or question prior, so it will be further fine-tuned in Section~\ref{subsection:3_5}.

\begin{figure}
   \begin{minipage}[b]{1.0\linewidth}
   \centerline{\includegraphics[width=85mm]{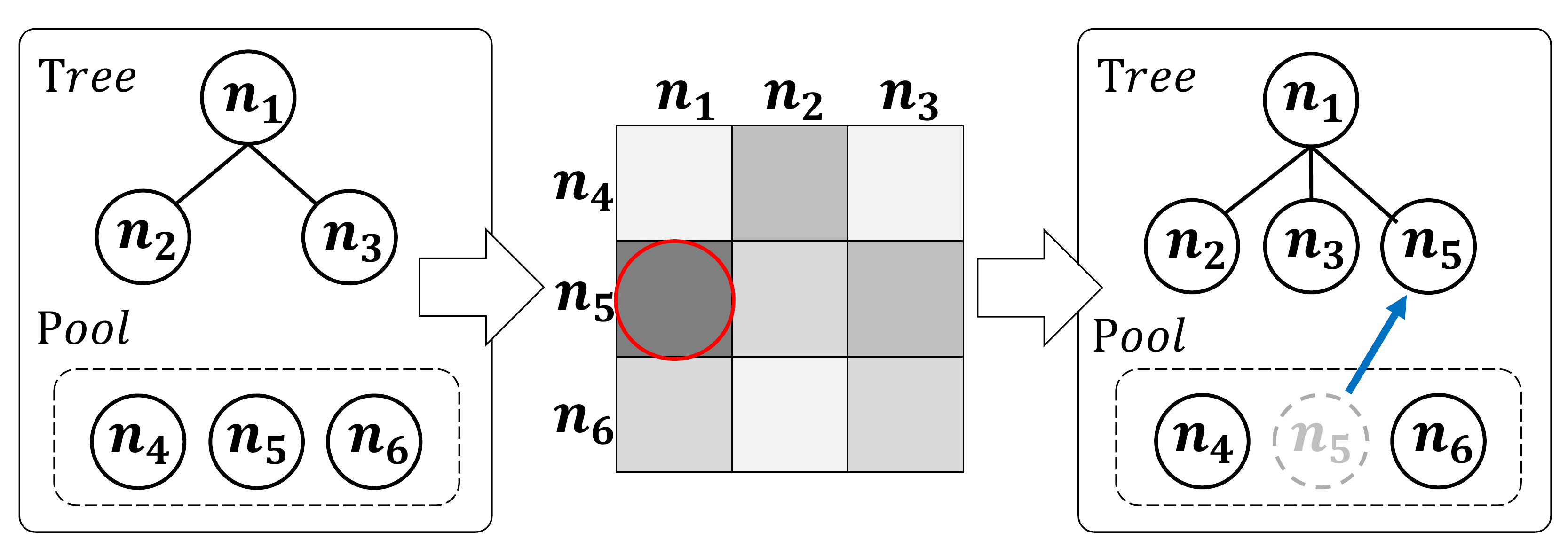}}
   \end{minipage}
   \caption{The maximum spanning tree from $\bm{S}$. In each step, a node in the remaining pool is connected to the current tree, if it has the highest validity score.}
   \vspace{-0.15in}
  \label{fig:3} 
\end{figure}

Considering $\bm{S}$ as a symmetric adjacency matrix, we can obtain a maximum spanning tree using the Prim's algorithm~\cite{prim1957shortest}, with a root (source node) $i$ satisfying $\arg\max_{i}\sum_{j~\ne~i}{\bm{S}_{ij}}$. In a nutshell, as illustrated in Figure~\ref{fig:3}, we construct the tree recursively by connecting the node from the pool to the tree node if it has the most validity. Note that during the tree structure exploration in Section~\ref{subsection:3_5}, each of the $i$-th step $t^{(i)}$ in the above tree construction is sampled from all possible choices in a multinomial distribution with the probability $p(t^{(i)} | t^{(1)},...,t^{(i-1)},\bm{S})$ in proportion to the validity. The resultant tree is multi-branch and is merely a sparse graph with only one kind of connection, which is still unable to discriminate the hierarchical and parallel relations in the subsequent context encoding. To this end, we convert the multi-branch tree into an equivalent binary tree, \ie, \textsc{VCTree} by changing non-leftmost edges into right branches as in Figure~\ref{fig:1}. In this fashion, the right branches (blue) indicate parallel contexts, and left ones (red) indicate hierarchical contexts. Such a binary tree structure achieves significant improvements in our SGG and VQA experiments compared to its multi-branch alternative.

\subsection{TreeLSTM Context Encoding}
\label{subsection:3_2}
Given the above constructed \textsc{VCTree}, we adopt BiTreeLSTM as our context encoder:
\begin{equation}
D = \textnormal{BiTreeLSTM}(\{\bm{z}_i\}_{i=1,2,...,n}),
\label{eq:2}
\end{equation}
where $\bm{z}_i$ is the input node feature, which will be specified in each task, and $D=[\bm{d}_1,\bm{d}_2,...,\bm{d}_n]$ is the encoded object-level visual context. Each $\bm{d}_i=[\vec{\bm{h}}_i;\cev{\bm{h}}_i]$ is the concatenated hidden states from both TreeLSTM~\cite{tai2015improved} directions:
\begin{align}
&\vec{\bm{h}}_i = \textnormal{TreeLSTM}(\bm{z}_i,\vec{\bm{h}}_{p}), \label{eq:3}\\
&\cev{\bm{h}}_i = \textnormal{TreeLSTM}(\bm{z}_i,[\cev{\bm{h}}_{l};\cev{\bm{h}}_{r}]),
\label{eq:4}
\end{align}
where $\vec{} \textnormal{and} \cev{}$ denote the top-down and bottom-up directions, respectively; we slightly abuse the subscripts $p,l,r$ to denote the parent, left child, and right child of node $i$. The order of the concatenation $[\cev{\bm{h}}_{l};\cev{\bm{h}}_{r}]$ in Eq.~\eqref{eq:4} indicates the explicit discrimination between the left and right branches in context encoding. We use zero vectors to pad all the missing branches.


\subsection{Scene Graph Generation Model}
\label{subsection:3_3}

\begin{figure}
   \begin{minipage}[b]{1.0\linewidth}
   \centerline{\includegraphics[width=85mm]{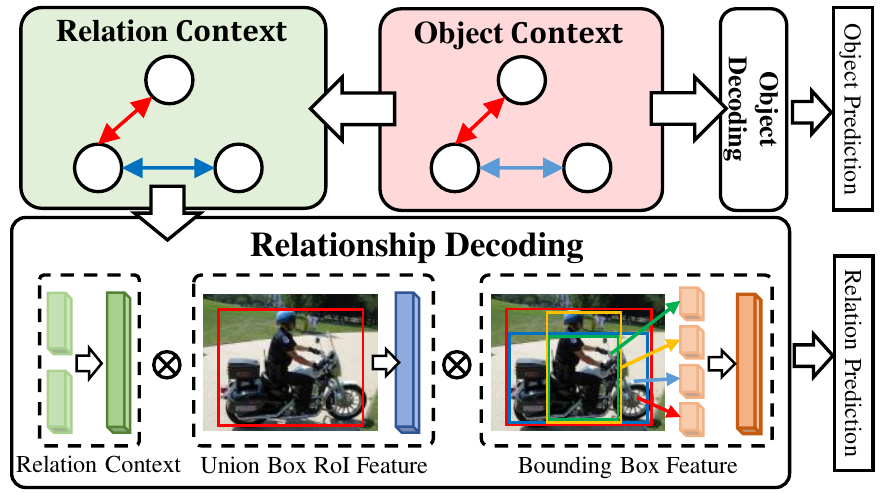}}
   \end{minipage}
   \caption{The overview of our SGG Model. The object context feature will be used to decode object categories, and the pairwise relationship decoding jointly fuses the relation context feature, RoIAlign feature of union box, and bounding box feature, before prediction.}
   \vspace{-0.1in}
   \label{fig:4} 
\end{figure}
Now we detail the implementation of Eq.~\eqref{eq:2} and how to decode them for the SGG task as illustrated in Figure~\ref{fig:4}. 

\noindent\textbf{Object Context Encoding}. We employ BiTreeLSTM from Eq.~\eqref{eq:2} to encode object context representation into $D^o=[\bm{d}_1^o,\bm{d}_2^o,...,\bm{d}_n^o], \bm{d}_i^o \in\mathbb{R}^{512}$. We set inputs $\bm{z}_i$ of Eq.~\eqref{eq:2} to $[\bm{x}_i;\bm{W}_1 \bm{\hat{c}}_i]$, \ie, concatenation of object visual features and embedded N-way original Faster-RCNN class probabilities, where $\bm{W}_1$ is the embedding matrix that maps each original label distribution $\bm{\hat{c}}_i$ into $\mathbb{R}^{200}$.

\noindent\textbf{Relation Context Encoding}. We apply an additional BiTreeLSTM using the above $\bm{d}_i^o$ as input $\bm{z}_i$ to further encode the relation context $D^r=[\bm{d}_1^r,\bm{d}_2^r,...,\bm{d}_n^r], \bm{d}_i^r \in\mathbb{R}^{512}$. 

\noindent\textbf{Context Decoding}. The goal of SGG is to detect objects and then predict their relationship. Similar to~\cite{zellers2017neural}, we adopt a dynamic object prediction which can be viewed as a decoding process in a top-down direction using Eq.~\eqref{eq:3}, that is, the object class of a child is dependent on its parent. Specifically, we set the input $\bm{z}_i$ of Eq.~\eqref{eq:3} to be $[\bm{d}_i^o;\bm{W}_2 \bm{c}_p]$, where $\bm{c}_p$ is the predicted label distribution of the $i$'s parent, and $\bm{W}_2$ embeds it into $\mathbb{R}^{200}$, then the output hidden is passed to a softmax classifier to achieve object label distribution $\bm{c}_i$. 

The relationship prediction is in a pairwise fashion. First, we collect three pairwise features for each object pair: (1) $\bm{d}_{ij} = \textnormal{MLP}([\bm{d}_i^r;\bm{d}_j^r])$ as the context feature, (2) $\bm{b}_{ij} = \textnormal{MLP}([\bm{b}_i; \bm{b}_j; \bm{b}_{i\cup j}; \bm{b}_{i\cap j}])$ as the bounding box pair feature, with $i\cup j, i\cap j$ being union box and intersection box, (3) $\bm{v}_{ij}$ as the RoIAlign feature~\cite{he2017mask} from the union bounding box of the object pair. All $\bm{d}_{ij}, \bm{v}_{ij}, \bm{b}_{ij}$ are under the same dimension $\mathbb{R}^{2048}$. Then, we fuse them into a final pairwise feature: $\bm{g}_{ij} = \bm{d}_{ij}\cdot \bm{v}_{ij}\cdot \bm{b}_{ij}$, before feed it into the softmax predicate classifier, where $\cdot$ is element-wise product.

\subsection{Visual Question Answering Model}
\label{subsection:3_4}
\begin{figure}
   \begin{minipage}[b]{1.0\linewidth}
   \centerline{\includegraphics[width=90mm]{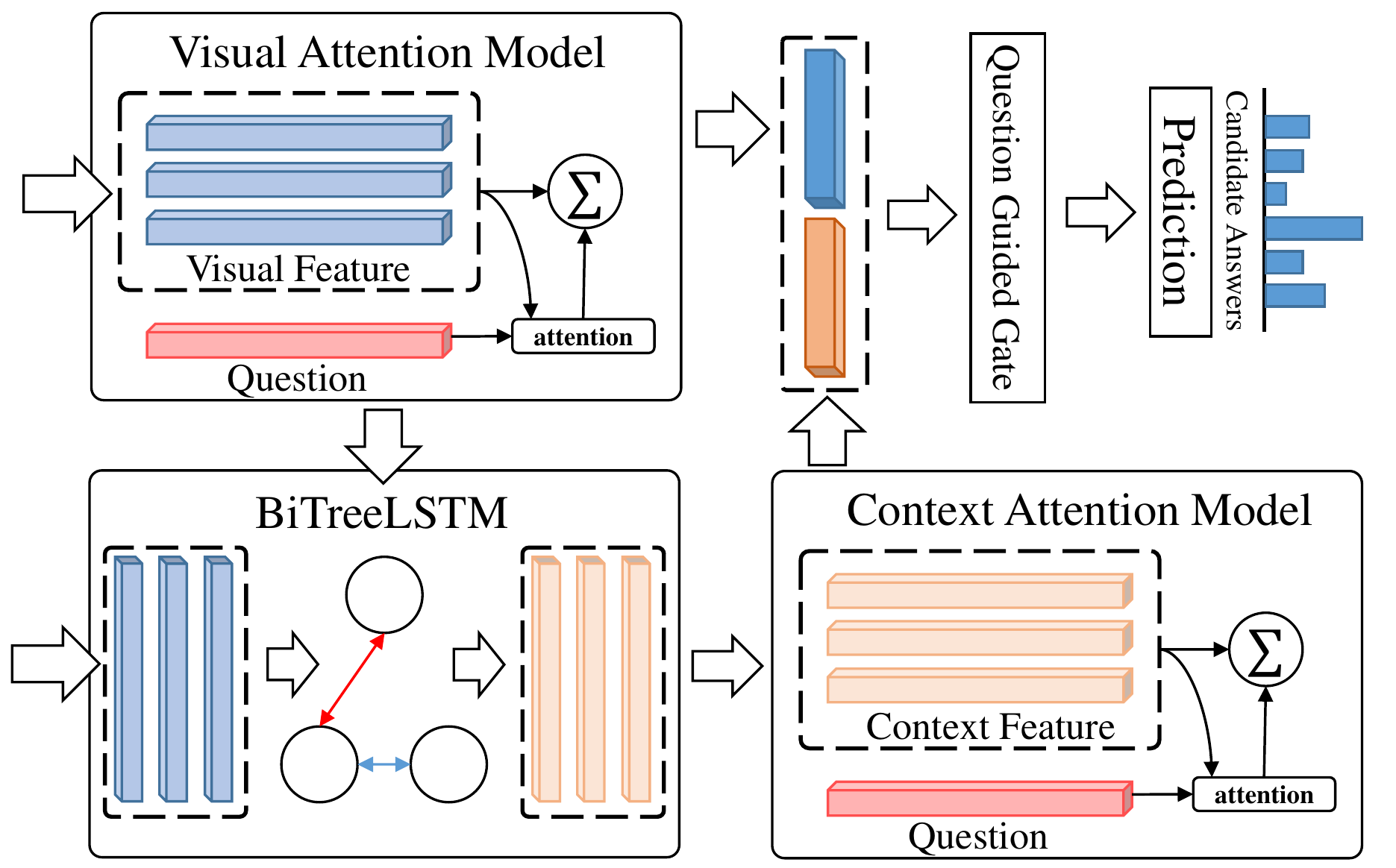}}
   \end{minipage}
   \caption{The overview of our VQA framework. It contains two multimodal attention models for visual feature and context feature. Outputs from both models will be concatenated and passed to a question-guided gate before answer prediction.}
   \vspace{-0.1in}
   \label{fig:5} 
\end{figure}

Now we detail the implementation of Eq.~\eqref{eq:2} for VQA, and illustrate our VQA model in Figure~\ref{fig:5}.

\noindent\textbf{Context Encoding}. The context feature in VQA: $D^q=[\bm{d}_1^q,\bm{d}_2^q,...,\bm{d}_n^q]$, $\bm{d}_i^q\in\mathbb{R}^{1024}$ is directly encoded from the bounding box visual feature $\bm{x}_i$ by Eq.~\eqref{eq:2}.

\noindent\textbf{Multimodal Attention Feature}. We adopt a popular attention model from previous work~\cite{anderson2018bottom, Teney_2018_CVPR} to calculate the multimodal joint feature $\bm{m}\in\mathbb{R}^{1024}$ for each question and image pair:
\begin{equation}
    \bm{m} = f_d(\bm{\hat{z}}, \bm{q}),
    \label{eq:8}
\end{equation}
where $\bm{q}\in\mathbb{R}^{1024}$ is the question feature from a one-layer GRU encoding the sentence; $\bm{\hat{z}} = \sum_{i=1}^N{\alpha_i \bm{z}_i}$ is the attentive image feature calculated from the input feature set $\{\bm{z}_i\}$, $\alpha_i = \exp{(u_i)}/\sum_k{\exp{(u_k)}}$ is the attention weight from object-task correlation $u_i=h(\bm{z}_i, \bm{q})=\textnormal{MLP}\big(f_d(\bm{z}_i, \bm{q})\big)$, with the output of MLP being a scalar; $f_d$ can be any multi-modal feature fusion function, in particular,  we adopt $f_d(\bm{x},\bm{y})=\textnormal{ReLU}(\bm{W}_3 \bm{x}+\bm{W}_4 \bm{y})-(\bm{W}_3 \bm{x}-\bm{W}_4 \bm{y})^2$ as in~\cite{zhang2018learning}, with $\bm{W}_3$ and $\bm{W}_4$ projecting $\bm{x}, \bm{y}$ into the same dimension. Therefore, we can use Eq.~\eqref{eq:8} to obtain both the multimodal visual attention feature $\bm{m}_x$ by setting input $\bm{z}_i$ to $\bm{x}_i$ and multimodal contextual attention feature $\bm{m}_d$ by setting $\bm{z}_i$ to $\bm{d}^q_i$.

\noindent\textbf{Question Guided Gate Decoding}. However, the importance of $\bm{m}_x$ and $\bm{m}_d$ varies from question to question, \eg, ``is there a dog?'' only requires visual features for detection, while ``is the man dressed formally?'' is highly context dependent. Inspired by~\cite{Shi_2018_ECCV}, we adopt a question guided gate to select the most related channels from $[\bm{m}_x;\bm{m}_d]$. The gate vector $\bm{g}\in\mathbb{R}^{2048}$ is defined as:
\begin{equation}
\bm{g} = \sigma\big(\textnormal{MLP}([\bm{q};\bm{W}_5 \bm{l}_q])\big),
\label{eq:12}
\end{equation}
where $\bm{l}_q\in\mathbb{R}^{65}$ is a one-hot question type vector defined by prefixed words of questions, which is embedded into $\mathbb{R}^{256}$ by matrix $\bm{W}_5$, and $\sigma(\cdot)$ denotes the sigmoid function. 

Finally, we fuse $\bm{g}\cdot [\bm{m}_x;\bm{m}_d]$ as the final VQA feature and feed it into the softmax classifier.

\begin{table*}[htbp]
\centering
\scalebox{.9}{
\begin{tabular}{l c c c|c c c|c c c}
\hline
\multicolumn{1}{c}{} & \multicolumn{3}{c}{Scene Graph Generation} & \multicolumn{3}{c}{Scene Graph Classification} & \multicolumn{3}{c}{Predicate Classification} \\
\hline
Model  & R@20 & R@50 & R@100 & R@20 & R@50 & R@100 & R@20 & R@50 & R@100 \\ 
\hline 
VRD~\cite{lu2016visual} & - & 0.3 & 0.5 & - & 11.8 & 14.1 & - & 27.9 & 35.0 \\
AsscEmbed~\cite{newell2017pixels} & 6.5 & 8.1 & 8.2 & 18.2 & 21.8 & 22.6 & 47.9 & 54.1 & 55.4 \\
IMP$^\diamond$~\cite{xu2017scene} & 14.6 & 20.7 & 24.5 & 31.7 & 34.6 & 35.4 & 52.7 & 59.3 & 61.3 \\
TFR~\cite{jae2018tensorize} & 3.4 & 4.8 & 6.0 & 19.6 & 24.3 & 26.6 & 40.1 & 51.9 & 58.3 \\
FREQ$^\diamond$~\cite{zellers2017neural} & 20.1 & 26.2 & 30.1 & 29.3 & 32.3 & 32.9 & 53.6 & 60.6 & 62.2 \\
MOTIFS$^\diamond$~\cite{zellers2017neural} & 21.4 & 27.2 & 30.3 & 32.9 & 35.8 & 36.5 & 58.5 & 65.2 & 67.1 \\
Graph-RCNN~\cite{Yang_2018_ECCV} & - & 11.4 & 13.7 & - & 29.6 & 31.6 & - & 54.2 & 59.1 \\
\hline 
Chain & 21.2 & 27.1 & 30.3 & 33.3 & 36.1 & 36.8 & 59.4 & 66.0 & 67.7 \\
Overlap & 21.4 & 27.3 & 30.4 & 33.7 & 36.5 & 37.1 & 59.5 & 66.0 & 67.8 \\
Multi-Branch & 21.5 & 27.3 & 30.6 & 34.3 & 37.1 & 37.8 & 59.5 & 66.1 & 67.8\\
\textsc{VCTree}-SL & 21.7 & 27.7 & 31.1 & 35.0 & 37.9 & 38.6 & 59.8 & 66.2 & 67.9 \\
\textsc{VCTree}-HL & \textbf{22.0} & \textbf{27.9} & \textbf{31.3} & \textbf{35.2} & \textbf{38.1} & \textbf{38.8} & \textbf{60.1} & \textbf{66.4} & \textbf{68.1} \\  
\hline
\hline
\end{tabular}
}
\caption{SGG performances (\%) of various methods. $^\diamond$ denotes the methods using the same Faster-RCNN detector as ours. IMP$^\diamond$ is reported from the re-implemented version~\cite{zellers2017neural}.}
\label{tab:1}
\end{table*}

\subsection{Hybrid Learning}
\label{subsection:3_5}
Due to the discrete nature of \textsc{VCTree} construction, the score matrix $\bm{S}$ is not fully differentiable from the loss back-propagated from the end-task loss. Inspired by~\cite{hu2017learning}, we use a hybrid learning strategy that combines reinforcement learning, \ie, policy gradient~\cite{williams1992simple} for the parameters $\theta$ of $\bm{S}$ in the tree construction and supervised learning for the rest parameters. Suppose a layout $l$, \ie, a constructed \textsc{VCTree}, is sampled from $\pi(l|I,q;\theta)$, \ie, the construction procedure in Section~\ref{subsection:3_1}, where $I$ is the given image, $q$ is the task, \eg, questions in VQA. To avoid clutter, we drop $I$ and $q$. Then, we define the reinforcement learning loss $L_{r}(\theta)$ as:
\begin{equation}
L_{r}(\theta)=-E_{l\sim\pi(l|\theta)}[r(l)],
\label{eq:13}
\end{equation}
where $L_{r}(\theta)$ aims to minimize the negative expected reward $r(l)$, which can be the end-task evaluation metrics such as Recall@100 for SGG and Accuracy for VQA. Then, the above gradient will be $\nabla_{\theta} L_r(\theta)=-E_{l\sim\pi(l|\theta)}[r(l)\nabla_\theta log\pi(l|\theta)]$. Since it is impractical to estimate all possible layouts, we use the Monte-Carlo sampling to estimate the gradient: 
\begin{equation}
\nabla_\theta L_r(\theta) \approx -\frac{1}{M} \sum_{m=1}^M \Big(r(l_m)\nabla_\theta log\pi(l_m|\theta)\Big),
\label{eq:14}
\end{equation}
where we set M to 1 in our implementation.

To reduce the gradient variance, we apply a self-critic baseline~\cite{Rennie_2017_CVPR} $b=r(\hat{l})$, where $\hat{l}$ is the greedy constructed tree without sampling. So the original reward $r(l_m)$ can be replaced by $r(l_m) - b$ in Eq.~\eqref{eq:14}. We observe faster convergence than using a traditional moving baseline~\cite{mnih2014recurrent}. 

The overall hybrid learning will be alternatively conducted between supervised learning and reinforcement learning, where we first train the supervised end-task on pretrained $\pi(l|\theta)$, then fix the end-task as reward function to learn our reinforcement policy network, after that, we update the supervised end-task by new $\pi(l|\theta)$. The latter two stages are running alternatively 2 times in our model.

\section{Experiments on Scene Graph Generation}
\subsection{Settings}
\noindent\textbf{Dataset.}
Visual Genome (VG)~\cite{krishna2017visual} is a popular benchmark for SGG. It contains 108,077 images with tens of thousands of unique object and predicate relation categories, yet most of categories have very limited instances. Therefore, previous works~\cite{li2017scene,xu2017scene,zhang2017relationship} proposed various VG splits that remove rare categories. We adopted the most popular one from~\cite{xu2017scene}, which selects top-150 object categories and top-50 predicate categories by frequency. The entire dataset is divided into the training set and test set by 70\%, 30\%, respectively. We further picked 5,000 images from training set as the validation set for hyper-parameter tuning.

\noindent\textbf{Protocols.}
We followed three conventional protocols to evaluate our SGG model: (1) \textbf{Scene Graph Generation (SGGen)}: given an image, detect object bounding boxes and their categories, and predict their relationships; (2) \textbf{Scene Graph Classification (SGCls)}: given ground-truth object bounding boxes in an image, predict the object categories and their relationships; (3) \textbf{Predicate Classification (PredCls)}: given the object categories and their bounding boxes in the image, predict their relationships. 

\noindent\textbf{Metrics.}
Since the annotation in VG is incomplete and biased, we followed the conventional Recall@K (R@K = 20,50,100) as the evaluation metrics~\cite{lu2016visual,xu2017scene,zellers2017neural}. However, it is well-known that SGG models trained on biased datasets such as VG have low performances for less frequent categories. To this end, we introduced a balanced metric called: \textbf{Mean Recall (mR@K)}. It calculates the recall on each predicate category independently, and then averages the results. So, each category contributes equally. Such a metric reduces the influence of some common yet meaningless predicates, \eg, ``on'', ``of'', and gives equal attention to those infrequent predicates, \eg, ``riding'', ``carrying'', which are more valuable to high-level reasoning. 

\begin{table}
\centering
\scalebox{0.9}
{
\begin{tabular}{l c | c | c}
\hline
\multicolumn{1}{c}{} & \multicolumn{1}{c}{SGGen} & \multicolumn{1}{c}{SGCls} & \multicolumn{1}{c}{PredCls} \\
\hline
Model & mR@100 & mR@100 & mR@100 \\ 
\hline 
MOTIFS$^\diamond$~\cite{zellers2017neural} & 6.6 & 8.2 & 15.3 \\
FREQ$^\diamond$~\cite{zellers2017neural}  & 7.1 & 8.5 & 16.0 \\
\textsc{VCTree}-HL & \textbf{8.0} & \textbf{10.8} & \textbf{19.4} \\  
\hline
\hline
\end{tabular}
}
\caption{Mean recall (\%) of various methods across all the 50 predicate categories.}
\vspace{-0.1in}
\label{tab:2}
\end{table}

\subsection{Implementation Details}
We adopted Faster-RCNN~\cite{ren2015faster} with VGG backbone to detect object bounding boxes and extract RoI features. Since the performance of SGG highly depends on the underlying detector, we used the same set of parameters as~\cite{zellers2017neural} for fair comparison. Object correlations $f(\bm{x}_i,\bm{x}_j)$ in Eq.~\eqref{eq:1} will be pretrained on ground-truth bounding boxes with class-agnostic relationships (\ie, foreground/background relationships), using all possible symmetric pairs without sampling. In SGGen, top-64 object proposals were selected after non-maximal suppression (NMS) with 0.3 IoU. We set background/foreground ratio for predicate classification to 3, and capped the number of training samples at 64 (retained all foreground pairs if possible). Our model is optimized by SGD with momentum, using learning rate $lr=6\cdot 10^{-3}$ and batch size $b=5$ for supervised learning, and $lr=6\cdot 10^{-4}, b=1$ for reinforcement learning. 

\begin{figure}
   \begin{minipage}[b]{1.0\linewidth}
   \centerline{\includegraphics[width=90mm]{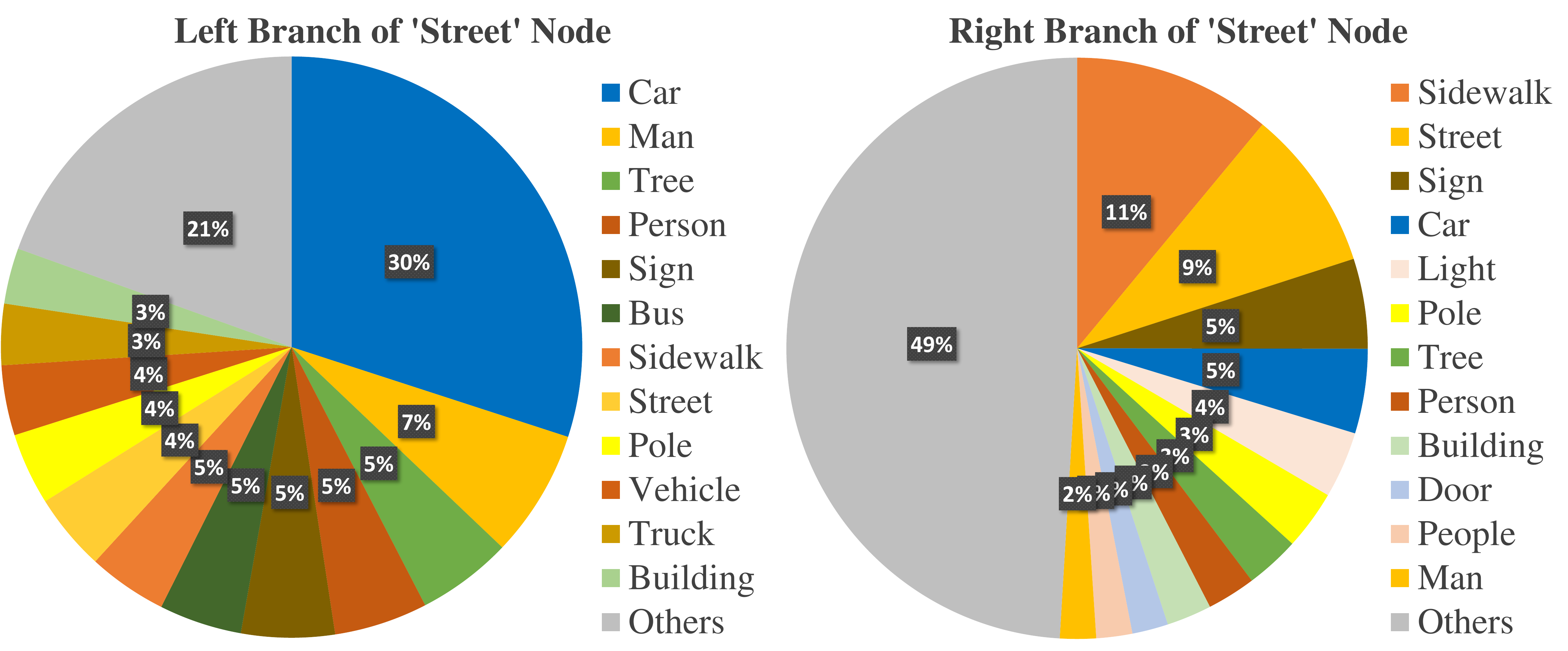}}
   \end{minipage}
   \caption{The statistics of left-branch (hierarchical) nodes and right-branch (parallel) nodes of the ``street'' category.}
   \vspace{-0.1in}
   \label{fig:6} 
\end{figure}

\subsection{Ablation Studies}
\label{subsection:4_4}
We investigated the influence of different structure construction policies. They are reported on the bottom half of Table~\ref{tab:1}. The ablative methods are (1) \textbf{Chain}: sorting all the objects by $\sum_{j:j\ne i}\bm{S}_{ij}$, then constructing a chain, which is different from the left-to-right ordered chain in MOTIFS~\cite{zellers2017neural}; (2) \textbf{Overlap}: iteratively constructing a binary tree by selecting the node with largest number of overlapped objects as parent, and dividing the rest nodes into left/right sub-trees by relatively positions of their bounding boxes; (3) \textbf{Multi-Branch}: the maximum spanning tree generated from score matrix $\bm{S}$, using Child-Sum TreeLSTM~\cite{tai2015improved} to incorporate context; (4) \textbf{\textsc{VCTree}-SL}: the proposed \textsc{VCTree} trained by supervised learning; (5) \textbf{\textsc{VCTree}-HL}: the complete version of \textsc{VCTree}, trained by hybrid learning for structure exploration in Section~\ref{subsection:3_5}. As we will show that Multi-Branch is significantly worse than \textsc{VCTree}, so there is no need to conduct hybrid learning experiment on Multi-Branch. We observe that \textsc{VCTree} performs better than other structures, and it is further improved by hybrid learning for structure exploration.

\begin{table}
\centering
\scalebox{0.8}{
\begin{tabular}{l c c c|c|c}
\hline
\multicolumn{6}{c}{VQA2.0 Validation Accuracy}\\
\hline
Model  & Yes/No & Number & Other & All & Balanced Pairs\\ 
\hline 
Graph & 81.8 & 44.9 & 56.6 & 64.5 & 36.3\\
Chain & 81.8 & 44.5 & 56.9 & 64.6 & 36.3\\
Overlap & 81.8 & 44.8 & 57.0 & 64.7 & 36.4\\ 
Multi-Branch & 82.1 & 44.3 & 56.9 & 64.7 & 36.6\\
\textsc{VCTree}-SL & 82.3 & 45.0 & 57.0 & 64.9 & 36.9\\
\textsc{VCTree}-HL & \textbf{82.6} & \textbf{45.1} & \textbf{57.1} & \textbf{65.1} & \textbf{37.2}\\ 
\hline
\hline
\end{tabular}
}
\caption{Accuracies (\%) of various context structures on the VQA2.0 validation set.}
\vspace{-0.1in}
\label{tab:3}
\end{table}

\subsection{Comparisons with State-of-the-Arts}
\noindent\textbf{Comparing Methods.} We compared \textsc{VCTree} with state-of-the-art methods in Table~\ref{tab:1}: (1) \textbf{VRD}~\cite{lu2016visual}, \textbf{FREQ}~\cite{zellers2017neural} are methods without using visual contexts. (2) \textbf{AssocEmbed}~\cite{newell2017pixels} assembles
implicit contextual features by stacked hourglass backbone~\cite{newell2016stacked}. (3) \textbf{IMP}~\cite{xu2017scene}, \textbf{TFR}~\cite{jae2018tensorize}, \textbf{MOTIFS}~\cite{zellers2017neural}, \textbf{Graph-RCNN}~\cite{Yang_2018_ECCV} are explicit context models with a variety of structures.

\noindent\textbf{Quantitative Analysis.} From Table~\ref{tab:1}, compared with the previous state-of-the-art MOTIFS~\cite{zellers2017neural}, the proposed \textsc{VCTree} has the best performances. Interestingly, Overlap tree and Multi-Branch tree are better than other non-tree context models. From Table~\ref{tab:2}, the proposed \textsc{VCTree}-HL shows larger absolute gains of PredCls under mR@100, which indicates that our model learns non-trivial visual context, \ie, not merely class distribution bias as in FREQ and partially in MOTIFS. Note that MOTIFS~\cite{zellers2017neural} is even worse than its FREQ~\cite{zellers2017neural} baseline under mR@100.

\begin{figure*}[t]
   \begin{minipage}[b]{1.0\linewidth}
   \centerline{\includegraphics[width=185mm]{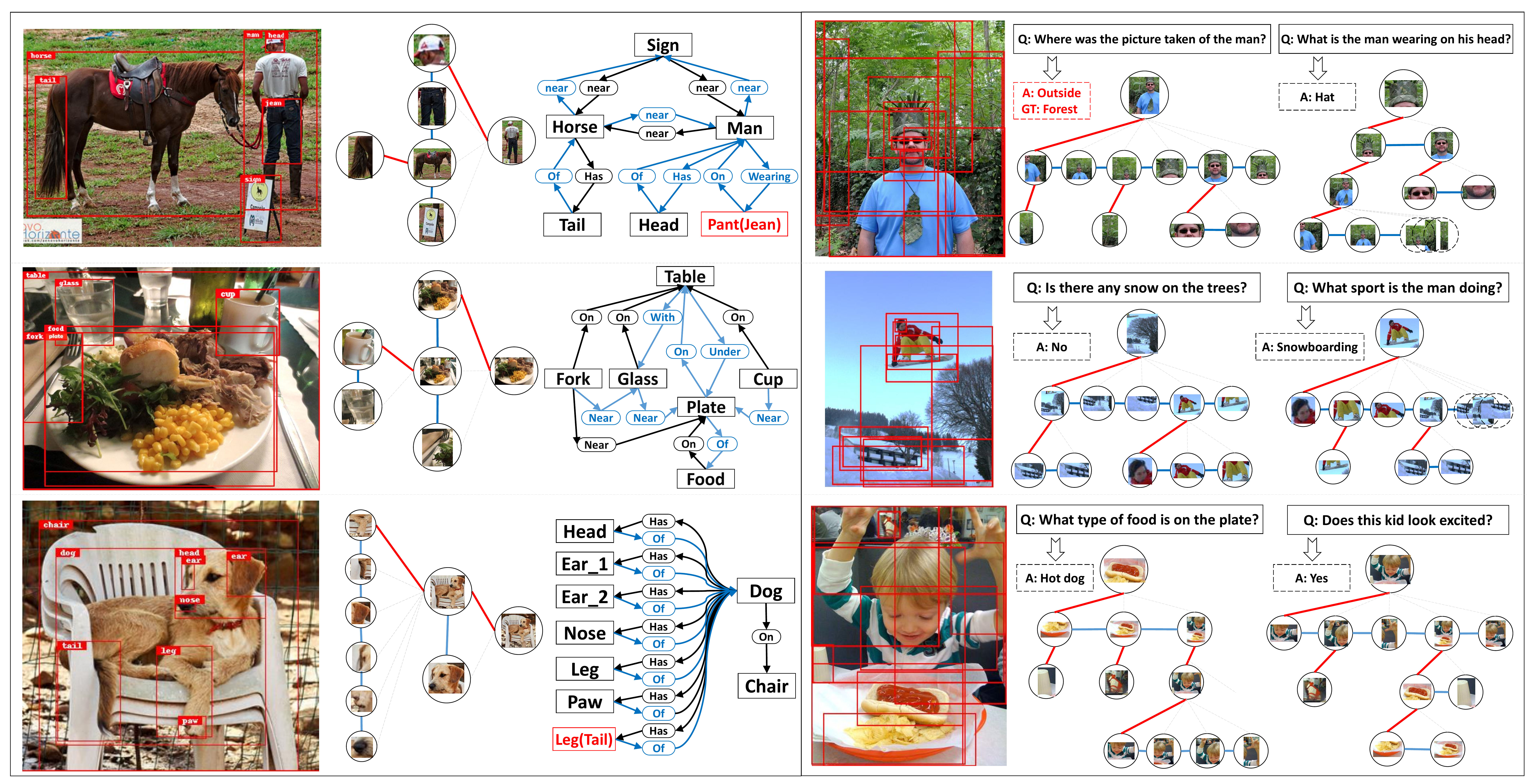}}
   \end{minipage}
   \caption{\textbf{Left:} the learned tree structure and generated scene graphs in VG. Black color indicates correctly detected objects or predicates; red indicates the misclassified ones; blue indicates correct predictions that not labeled as ground-truth. \textbf{Right:} interpretable and dynamic trees subject to different questions in VQA2.0.}
   \vspace{-0.1in}
   \label{fig:7} 
\end{figure*}

\noindent\textbf{Qualitative Analysis.}
To better understand what context is learned by \textsc{VCTree}, we visualized a statistics of left-/right-branch nodes for nodes classified as ``street'' in Figure~\ref{fig:6}. From the left pie, the hierarchical relations, we can see the node categories are long-tailed, \ie, top-10 categories cover the 73\% of the instances; while the right pie, the parallel relations, are more uniformly distributed. This demonstrates that \textsc{VCTree} captures the two types of context successfully. More qualitative examples of \textsc{VCTree}s and their generated scene graph can be viewed in Figure~\ref{fig:7}. The common errors are generally synonymous labels, \eg, ``jeans'' vs. ``pants'', ``man'' vs. ``person'', and over-interpretation, \eg, the ``tail'' of bottom left ``dog'' is considered as ``leg'', as it appears at the place where ``leg'' should be.



\begin{table}
\centering
\scalebox{.8}{
\begin{tabular}{l c c c|c}
\hline
\multicolumn{5}{c}{VQA2.0 test-dev} \\
\hline
Model  & Yes/No & Number & Other & All\\ 
\hline 
Teney~\cite{Teney_2018_CVPR} & 81.82 & 44.21 & 56.05 & 65.32\\
MUTAN~\cite{ben2017mutan} & 82.88 & 44.54 & 56.50 & 66.01 \\
MLB~\cite{kim2016hadamard} & 83.58 & 44.92 & 56.34 & 66.27 \\
DA-NTN~\cite{bai2018deep} & \textbf{84.29} & 47.14 & 57.92 & 67.56\\
Count~\cite{zhang2018learning} & 83.14 & \textbf{51.62} & 58.97 & 68.09\\ 
\hline
Chain & 82.74 & 47.31 & 58.93 & 67.42\\
Graph & 83.53 & 47.09 & 58.6 & 67.56\\
\textsc{VCTree}-HL & 84.28 & 47.78 & \textbf{59.11} & \textbf{68.19}\\
\hline
\hline
\end{tabular}
}
\caption{Single-model accuracies (\%) on VQA2.0 test-dev, where MUTAN and MLB are re-implemented versions from~\cite{bai2018deep}.}
\label{tab:4}
\end{table}

\begin{table}
\centering
\scalebox{.8}{
\begin{tabular}{l c c c|c}
\hline
\multicolumn{5}{c}{VQA2.0 test-standard} \\
\hline
Model  & Yes/No & Number & Other & All\\ 
\hline 
Teney~\cite{Teney_2018_CVPR} & 82.20 & 43.90 & 56.26 & 65.67 \\
MUTAN~\cite{ben2017mutan} & 83.06 & 44.28 & 56.91 & 66.38 \\
MLB~\cite{kim2016hadamard} & 83.96 & 44.77 & 56.52 & 66.62 \\
DA-NTN~\cite{bai2018deep} & \textbf{84.60} & 47.13 & 58.20 & 67.94 \\
Count~\cite{zhang2018learning} & 83.56 & \textbf{51.39} & 59.11 & 68.41 \\ 
\hline
Chain & 83.06 & 47.38 & 58.95 & 67.68\\
Graph & 84.03 & 47.08 & 58.82 & 68.0\\
\textsc{VCTree}-HL & 84.55 & 47.36 & \textbf{59.34} & \textbf{68.49} \\
\hline
\hline
\end{tabular}
}
\caption{Single-model accuracies (\%) on VQA2.0 test-standard, where MUTAN and MLB are re-implemented versions from \cite{bai2018deep}.}
\vspace{-0.1in}
\label{tab:5}
\end{table}

\section{Experiments on Visual Q\&A}
\subsection{Settings}
\noindent\textbf{Datasets.}
We evaluated the proposed VQA model on VQA2.0~\cite{goyal2017making}. Compared with VQA1.0~\cite{antol2015vqa}, VQA2.0 has more question-image pairs for training (443,757) and validation (214,354), and all the question-answer pairs are balanced by making sure the same question can have different answers. In VQA2.0, the ground-truth accuracy of a candidate answer is considered as the average of $\min(\frac{\#\textnormal{Humans~votes}}{3}, 1)$ over all 10 select 9 sets. Question-answer pairs are organized in three answer types:~\ie~``Yes/No'', ``Number'', ``Other''. There are also 65 question types determined by prefixed words, which we used to generate question-guided gates. We also tested our models on a balanced subset of validation set, called Balanced Pairs~\cite{Teney_2018_CVPR}, which requires the same question on different images with two different yet perfect (with 1.0 ground-truth score) answers. Since Balanced Pairs strictly removes question-related bias, it reflects the ability of a context model to distinguish subtle differences between images.

\subsection{Implementation Details}
We employed a simple text preprocessing for questions and answers, which changes all characters into lower-case and removes special characters. Questions were encoded into a vocabulary of the size 13,758 without trimming. Answers used a 3,000 vocabulary selected by frequency. For fair comparison, we used the same bottom-up feature~\cite{anderson2018bottom} as previous methods~\cite{anderson2018bottom, bai2018deep, Teney_2018_CVPR, zhang2018learning}, which contains 10 to 100 object proposals per image extracted by Faster-RCNN~\cite{ren2015faster}. We used the same Faster-RCNN detector to pretrain the $f(\bm{x}_i,\bm{x}_j)$. Since candidate answers were represented by probabilities rather than one-hot vectors in VQA2.0, we allowed the cross-entropy loss calculating soft categories, \ie, probabilities of ground-truth candidate answers. We used Adam optimizer with learning rate $lr=0.0015$ and batch size $b=256$, $lr$ decayed at ratio of 0.5 every 20 epochs. 


\subsection{Ablation Studies}
In addition to the 5 structure construction policies introduced in Section~\ref{subsection:4_4}, we also implemented a fully-connected graph structure using the message passing mechanism~\cite{xu2017scene}. From Table~\ref{tab:3}, the proposed \textsc{VCTree}-HL outperforms all the context models on three answer types. 

We further evaluated the above context models on VQA2.0 balanced pair subset~\cite{Teney_2018_CVPR}: the last column of Table~\ref{tab:3}, and found that the absolute gains between \textsc{VCTree}-HL and other structures are even larger than those on the original validation set. Meanwhile, as reported in~\cite{Teney_2018_CVPR}, different architectures or hyper-parameters in non-contextual VQA model normally gain less improvements on the balanced pair subset than overall validation set. Thus, it suggests that \textsc{VCTree} indeed use better context structures to alleviate the question-answer bias in VQA.

\subsection{Comparisons with State-of-the-Arts}
\noindent\textbf{Comparing Methods.} Table~\ref{tab:4}~\&~\ref{tab:5} reports the single-model performances of various state-of-the-art methods~\cite{bai2018deep, ben2017mutan, kim2016hadamard, Teney_2018_CVPR, zhang2018learning} on both test-dev and test-standard sets. For fair comparison, the reported methods are all using the same Faster-RCNN features~\cite{anderson2018bottom} as ours.

\noindent\textbf{Quantitative Analysis.} The proposed \textsc{VCTree}-HL shows the best overall performance in both test-dev and test-standard. Note that though Count~\cite{zhang2018learning} has close overall performance to our \textsc{VCTree}, it mainly improves the ``Number'' task by the elaborately designed model, while the proposed \textsc{VCTree} is a more general solution.  

\noindent\textbf{Qualitative Analysis.} We visualized several examples of \textsc{VCTree}-HL on the validation set. They illustrate that the proposed \textsc{VCTree} is able to learn dynamic structures with interpretability, \eg, in Figure~\ref{fig:7}, given the right middle image with the question ``Is there any snow on the trees?'', the generated \textsc{VCTree} locates the ``tree'' then searching for the ``snow'', while with question ``What sport is the man doing?'', the ``man'' appears to be the root.


\section{Conclusions}
In this paper, we proposed a dynamic tree structure called \textsc{VCTree} to capture task-specific visual contexts, which can be encoded to support two high-level vision tasks: SGG and VQA. By exploiting \textsc{VCTree}, we observed consistent performance gains in SGG on Visual Genome and in VQA on VQA2.0, compared to models with or without visual contexts. Besides, to justify that \textsc{VCTree} learns non-trivial contexts, we conducted additional experiments against the category bias in SGG and the question-answer bias in VQA, respectively. In the future, we intend to study the potential of a dynamic forest as the underlying context structure.

{\small
\bibliographystyle{ieee}
\bibliography{egpaper_final}
}

\newpage

\appendix
\section{Bidirectional TreeLSTM}
In this section, we will introduce the details of the bidirectional TreeLSTM applied to encode the object-level visual contexts. For the bottom-up direction, we employ $N$-ary TreeLSTM~\cite{tai2015improved} for binary trees, \ie, \textsc{VCTree}s and Overlap Trees, and the normalized Child-Sum~\cite{tai2015improved} TreeLSTM for Multi-Branch Trees. For the top-down direction, since each node only has one parent, TreeLSTM is similar to the traditional LSTM~\cite{hochreiter1997long}.

\subsection{$N$-ary TreeLSTM for Binary Trees}
According to the definition of $N$-ary TreeLSTM~\cite{tai2015improved}, it can be applied to the tree structures with at most $N$ ordered branches for each node. In our work, we adopt binary TreeLSTM as our bottom-up TreeLSTM for the proposed binary tree structures, \ie, \textsc{VCTree}s and Overlap Trees. It can be formulated as follows:
\begin{align}
\cev{\bm{h}}_t &= \textnormal{TreeLSTM}\big(\bm{z}_t,[\cev{\bm{h}}_{l};\cev{\bm{h}}_{r}]\big), \\
\bm{i}_t &= \sigma\big(\bm{W}^{(i)}\bm{z}_t + \bm{U}^{(i)}[\cev{\bm{h}}_{l};\cev{\bm{h}}_{r}] + \bm{b}^{(i)}\big), \\
\bm{f}_l &=  \sigma\big(\bm{W}^{(f)}_l\bm{z}_t + \bm{U}^{(f)}_l[\cev{\bm{h}}_{l};\cev{\bm{h}}_{r}] + \bm{b}_l^{(f)}\big), \\
\bm{f}_r &=  \sigma\big(\bm{W}^{(f)}_r\bm{z}_t + \bm{U}^{(f)}_r[\cev{\bm{h}}_{l};\cev{\bm{h}}_{r}] + \bm{b}_r^{(f)}\big), \\
\bm{o}_t &= \sigma\big(\bm{W}^{(o)}\bm{z}_t + \bm{U}^{(o)}[\cev{\bm{h}}_{l};\cev{\bm{h}}_{r}] + \bm{b}^{(o)}\big), \\
\bm{u}_t &= \textnormal{tanh}\big(\bm{W}^{(u)}\bm{z}_t + \bm{U}^{(u)}[\cev{\bm{h}}_{l};\cev{\bm{h}}_{r}] + \bm{b}^{(u)}\big), \\
\cev{\bm{c}}_t &= \bm{i}_t \odot \bm{u}_t + \bm{f}_l \odot \cev{\bm{c}}_l + \bm{f}_r \odot \cev{\bm{c}}_r, \\
\cev{\bm{h}}_t &= \bm{o}_t \odot \textnormal{tanh}(\cev{\bm{c}}_t),
\label{eq:s_1}
\end{align}
where $\bm{z}_t\in\mathbb{R}^d$ is the input feature for node $t$; $\cev{\bm{h}_t}, \cev{\bm{h}_l}, \cev{\bm{h}_r} \in \mathbb{R}^h$ are the hidden states; $\cev{\bm{c}_t}, \cev{\bm{c}_l}, \cev{\bm{c}_r} \in \mathbb{R}^h$ are memory cells; $\bm{W}^{(i)}, \bm{W}^{(f)}_l, \bm{W}^{(f)}_r, \bm{W}^{(o)}, \bm{W}^{(u)} \in \mathbb{R}^{h\times d}$ and $\bm{U}^{(i)}, \bm{U}^{(f)}_l, \bm{U}^{(f)}_r, \bm{U}^{(o)}, \bm{U}^{(u)} \in \mathbb{R}^{h\times 2h}$ are learnable matrices; $\bm{b}^{(i)}, \bm{b}^{(f)}_l, \bm{b}^{(f)}_r, \bm{b}^{(o)}, \bm{b}^{(u)} \in \mathbb{R}^{h}$ are vectors; $\sigma$ denotes sigmoid function; $\textnormal{tanh}$ denotes tanh activation function; $\odot$ means element-wise product. Note that we slightly abuse the subscripts $l,r$ of $\cev{\bm{c}}_l, \cev{\bm{c}}_r, \cev{\bm{h}}_l, \cev{\bm{h}}_r$ to denote hidden states and memory cells from the left-child and right-child of node $t$. The hidden states and memory cells of the missing branches will be filled with zero vectors.

\subsection{Child-Sum TreeLSTM for Multi-Branch Trees}
The Child-Sum TreeLSTM~\cite{tai2015improved} is able to deal with the tree structure where each node has arbitrary number of children. Therefore, we adopt it as the bottom-up TreeLSTM of the context encoder for the Multi-Branch Trees in the ablation studies. For each node $t$ of a Multi-Branch Tree, we define $C(t)$ as the set of its children. Compared with the original paper~\cite{tai2015improved}, we replace the Child-Sum with the Child-Mean in our implementation for better normalization, then it is formulated as:
\begin{align}
&\cev{\bm{h}}_t = \textnormal{TreeLSTM}\big(\bm{z}_t,\{\cev{\bm{h}}_{k}\}\big), k\in C(t), \\
&\cev{\bm{h}}_{mean} = \frac{ \sum_{k\in C(t)} \cev{\bm{h}}_k }{ |C(t)| }, \\
&\bm{i}_t = \sigma\big(\bm{W}^{(i)}\bm{z}_t + \bm{U}^{(i)}\cev{\bm{h}}_{mean} + \bm{b}^{(i)}\big), \\
&\bm{f}_k =  \sigma\big(\bm{W}^{(f)}\bm{z}_t + \bm{U}^{(f)}\cev{\bm{h}}_k + \bm{b}^{(f)}\big), \\
&\bm{o}_t = \sigma\big(\bm{W}^{(o)}\bm{z}_t + \bm{U}^{(o)}\cev{\bm{h}}_{mean} + \bm{b}^{(o)}\big), \\
&\bm{u}_t = \textnormal{tanh}\big(\bm{W}^{(u)}\bm{z}_t + \bm{U}^{(u)}\cev{\bm{h}}_{mean} + \bm{b}^{(u)}\big), \\
&\cev{\bm{c}}_t = \bm{i}_t \odot \bm{u}_t + \frac{ \sum_{k\in C(t)} \bm{f}_k \odot \cev{\bm{c}}_k }{ |C(t)|}, \\
&\cev{\bm{h}}_t = \bm{o}_t \odot \textnormal{tanh}(\cev{\bm{c}}_t),
\label{eq:s_2}
\end{align}
where $\cev{\bm{h}_t}, \cev{\bm{h}_k}, \in \mathbb{R}^h$ are the hidden states; $\cev{\bm{c}_t}, \cev{\bm{c}_k} \in \mathbb{R}^h$ are memory cells; $\bm{W}^{(i)}, \bm{W}^{(f)}, \bm{W}^{(o)}, \bm{W}^{(u)} \in \mathbb{R}^{h\times d}$ and $\bm{U}^{(i)}, \bm{U}^{(f)}, \bm{U}^{(o)}, \bm{U}^{(u)} \in \mathbb{R}^{h\times h}$ are learnable matrices; $\bm{b}^{(i)}, \bm{b}^{(f)}, \bm{b}^{(o)}, \bm{b}^{(u)} \in \mathbb{R}^{h}$ are vectors; $|C(t)|$ is the number of children for node $t$; $\cev{\bm{h}}_{mean}$ denotes the mean hidden state of all the children of node $t$.

\begin{table*}
\centering
\scalebox{.9}{
\begin{tabular}{l c c c|c c c|c c c}
\hline
\multicolumn{1}{c}{} & \multicolumn{3}{c}{Scene Graph Generation} & \multicolumn{3}{c}{Scene Graph Classification} & \multicolumn{3}{c}{Predicate Classification} \\
\hline
Model  & mR@20 & mR@50 & mR@100 & mR@20 & mR@50 & mR@100 & mR@20 & mR@50 & mR@100 \\ 
\hline 
MOTIFS~\cite{zellers2017neural} & 4.2 & 5.7 & 6.6 & 6.3 & 7.7 & 8.2 & 10.8 & 14.0 & 15.3 \\
FREQ~\cite{zellers2017neural} & 4.5 & 6.1 & 7.1 & 5.1 & 7.2 & 8.5 & 8.3 & 13.0 & 16.0 \\
Chain & 4.6 & 6.3 & 7.2 & 6.3 & 7.9 & 8.8 & 11.0 & 14.4 & 16.6 \\
Overlap & 4.8 & 6.5 & 7.5 & 7.2 & 9.0 & 9.3 & 12.5 & 16.1 & 17.4 \\
Multi-Branch & 4.7 & 6.5 & 7.4 & 6.9 & 8.6 & 9.2 & 11.9 & 15.5 & 16.9\\
\textsc{VCTree}-SL & 5.0 & 6.7 & 7.7 & 8.0 & 9.8 & 10.5 & 13.4 & 17.0 & 18.5 \\
\textsc{VCTree}-HL & \textbf{5.2} & \textbf{6.9} & \textbf{8.0} & \textbf{8.2} & \textbf{10.1} & \textbf{10.8} & \textbf{14.0} & \textbf{17.9} & \textbf{19.4} \\  
\hline
\hline
\end{tabular}
}
\caption{Mean recall (\%) of various methods across all the 50 predicate categories. MOTIFS~\cite{zellers2017neural} and FREQ~\cite{zellers2017neural} are using the same Faster-RCNN detector as ours.}
\label{tab:s_1}
\end{table*}

\begin{figure*}
   \begin{minipage}[b]{1.0\linewidth}
   \centerline{\includegraphics[width=180mm]{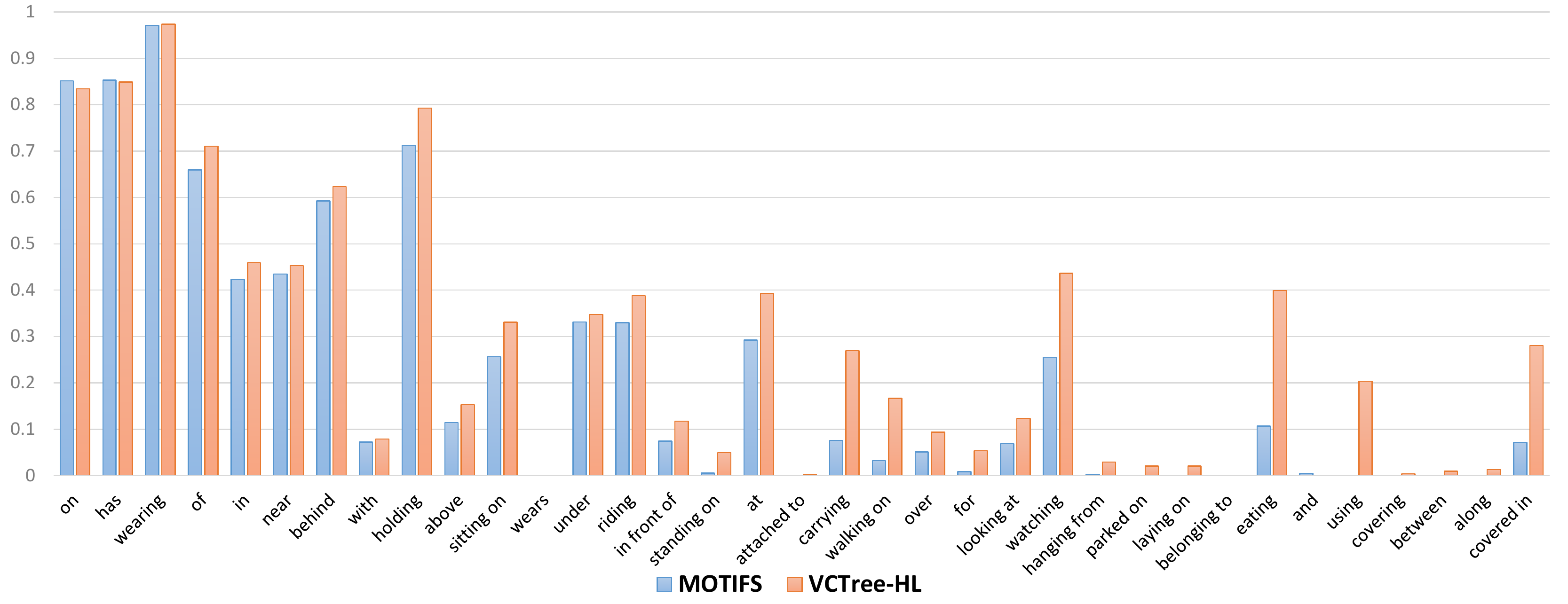}}
   \end{minipage}
   \caption{Recall@100 of MOTIFS~\cite{zellers2017neural} and the proposed \textsc{VCTree}-HL under PredCls for each Top-35 category ranking by frequency.}
  \label{fig:s_1} 
\end{figure*}

\subsection{Top-Down TreeLSTM}
We use the traditional LSTM~\cite{hochreiter1997long} as the top-down TreeLSTM for all the \textsc{VCTree}s, Overlap Trees, and Multi-Branch Trees, because each node only has at most one parent. The only difference with the traditional LSTM is that our structures are trees rather than chains, the previous hidden state is from the parent of node $t$. 

For the proposed \textsc{VCTree}, we assigned different learnable matrices for the hidden states from the left-branch parents and right-branch parents. However, the result didn't show significant improvements in the end-tasks, so we employ traditional LSTM as our top-down LSTM for efficiency.

\section{Quantitative Analysis}
\subsection{Mean Recall for Scene Graph}
We also report more detailed results of the proposed \textbf{Mean Recall (mR@K)} in Table~\ref{tab:s_1}. The proposed \textsc{VCTree}-HL shows best performance among all the ablative structures. Note that MOTIFS~\cite{zellers2017neural} has lower mR@100 than FREQ~\cite{zellers2017neural} baseline in SGCls and PredCls, which means that MOTIFS is even worse at predicting infrequent predicate categories. However, its mR@20 and mR@50 are higher than FREQ in SGCls and PredCls, which indicates that MOTIFS better separates the foreground relationships from the background ones than FREQ. 

\subsection{Predicate Recall Analysis}
To better visualize the improvement of the proposed \textsc{VCTree}-HL on infrequent predicate categories, we rank all the predicate categories by frequency, and show the PredCls Recall@100 of MOTIFS~\cite{zellers2017neural} and \textsc{VCTree}-HL for each top-35 category independently in Figure~\ref{fig:s_1}. We can observe significant improvements on those less frequent but more semantically meaningful predicates.  

\section{Qualitative Analysis}
\subsection{Scene Graph Generation}
\begin{figure*}
   \begin{minipage}[b]{1.0\linewidth}
   \centerline{\includegraphics[width=170mm]{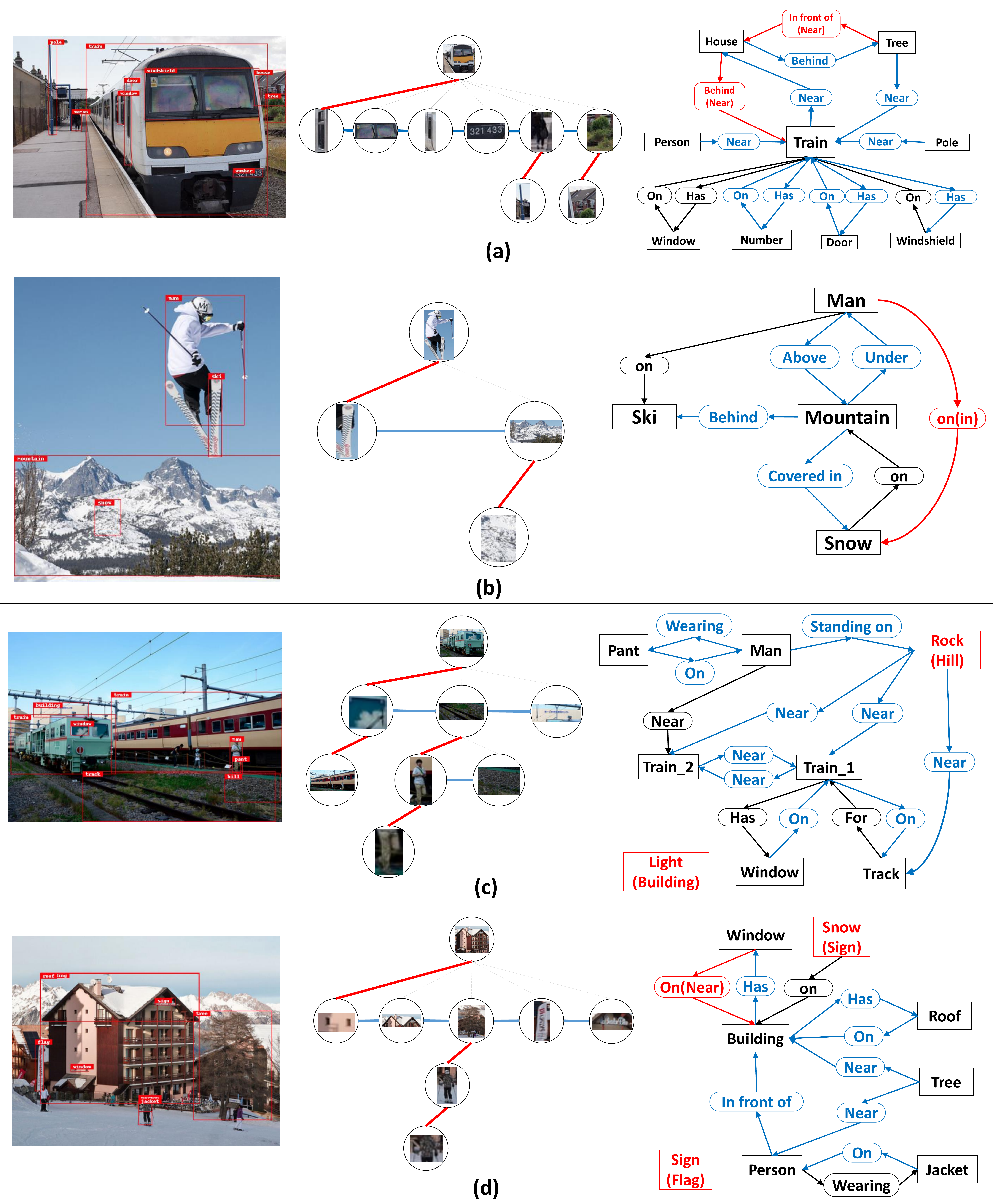}}
   \end{minipage}
   \caption{The learned tree structures and generated scene graphs in VG. We selectively report the predicates from R@20 and all the ground-truth predicates. Black color indicates correctly detected objects or predicates; red indicates the misclassified ones; blue indicates correct predictions that not labeled as ground-truth.}
  \label{fig:s_2} 
\end{figure*}

We further investigated more misclassified results of the proposed \textsc{VCTree}-HL. The corresponding tree structures and the generated scene graphs are reported in Figure~\ref{fig:s_2}. We observed 3 types of interesting misclassifications: 1) In the image (a) of Figure~\ref{fig:s_2}, the proposed \textsc{VCTree}-HL predicts more appropriate predicates ``in front of'' and ``behind'' than original ``near''. 2) In the image (b) and (d), the ground truth ``man in snow'' and ``window near building'' are improper, while our method shows more appropriate predicates. 3) In the image (c) and (d), the objects isolated from the Scene Graph (only considering R@20 predicates) are easier to be misclassified.

\subsection{Visual Question Answering}
\begin{figure*}
   \begin{minipage}[b]{1.0\linewidth}
   \centerline{\includegraphics[width=170mm]{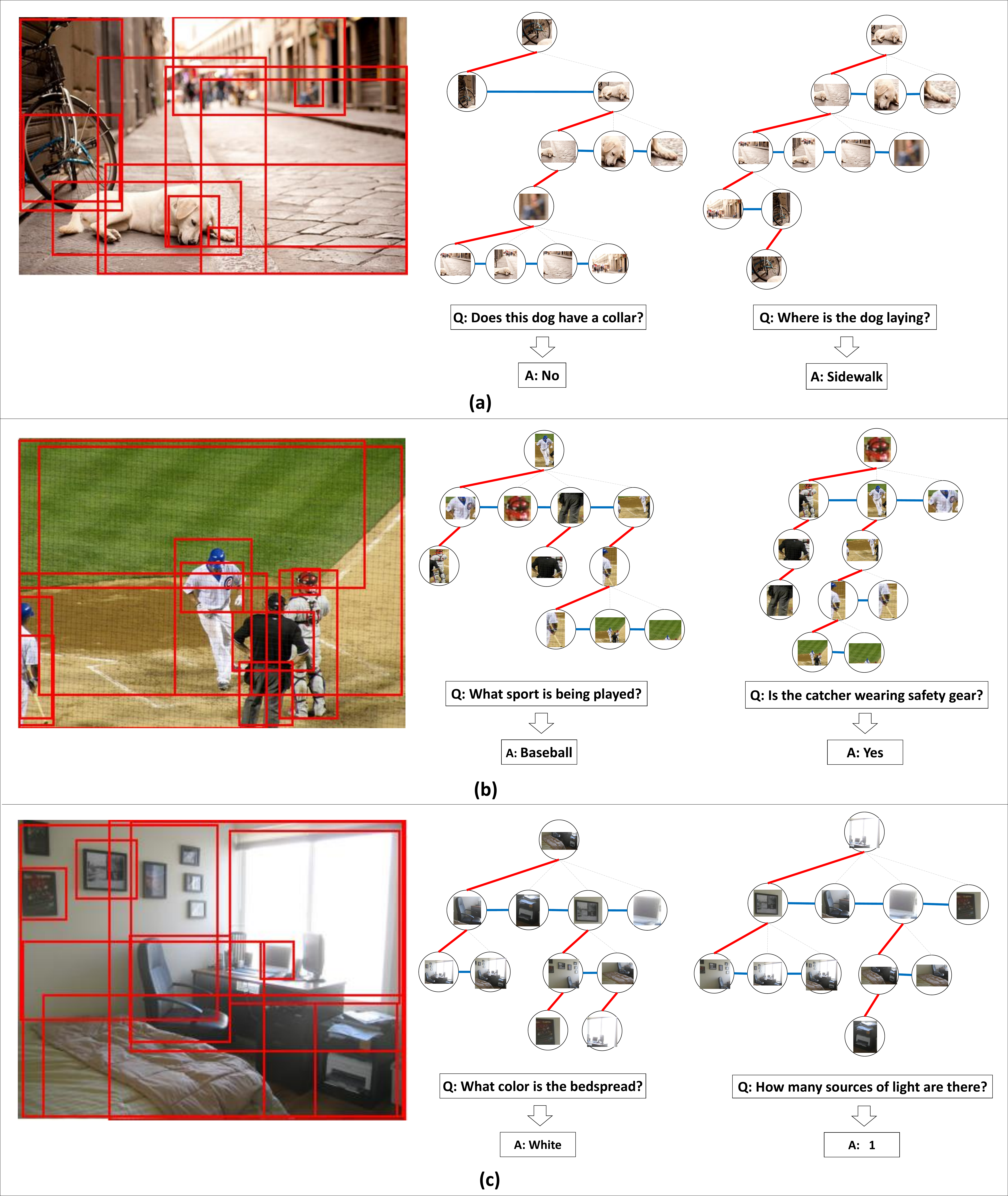}}
   \end{minipage}
   \caption{The dynamic and interpretable tree structures that subject to different questions, which allow the objects in an image incorporate different contextual cues according to each question.}
  \label{fig:s_3} 
\end{figure*}

More constructed \textsc{VCTree}s for VQA2.0 are visualized in Figure~\ref{fig:s_3}. The dynamic tree structures are subject to different questions, which allow the objects in an image to incorporate the different contextual cues according to each question. The proposed \textsc{VCTree} also helps us understand how the model predicts the answer of the question given the image, \eg, in image (a) of Figure~\ref{fig:s_3}, given the question ``does this dog have a collar?'', we find that our model first focuses on the collar-like object rather than the dog; in image (b) of Figure~\ref{fig:s_3}, given the question ``what sport is being played?'', we find that our model focuses on the sportsman rather than playground to answer this question.

\end{document}